\documentclass{article}
\usepackage{spconf,amsmath,graphicx}
\usepackage{xcolor}
\usepackage{amsmath,bbm}
\usepackage{amssymb}
\usepackage{booktabs}
\usepackage{multirow}
\usepackage{cite}
\usepackage[mathscr]{euscript}
\usepackage{fancyhdr}

\DeclareMathOperator*{\argmax}{arg\,max}

\DeclareMathAlphabet\mathbfcal{OMS}{cmsy}{b}{n}

\title{Back-and-Forth Prediction for Deep Tensor Compression}
\name{Hyomin Choi, Robert A. Cohen and Ivan V. Baji\'{c}}
\address{School of Engineering Science, Simon Fraser University, Burnaby, BC, Canada}

\begin{document}

\maketitle

\thispagestyle{empty}
\renewcommand{\headrulewidth}{0.0pt}
\thispagestyle{fancy}
\lhead{}
\chead{Copyright \copyright 2020 IEEE. Personal use of this material is permitted. However, permission to use this material for any other purposes must be obtained from the IEEE by sending an email to pubs-permissions@ieee.org.}
\rhead{}
\lfoot{}
\cfoot{}
\rfoot{}

\begin{abstract}

Recent AI applications such as Collaborative Intelligence with neural networks involve transferring deep feature tensors between various computing devices. This necessitates tensor compression in order to optimize the usage of bandwidth-constrained channels between devices. In this paper we present a prediction scheme called Back-and-Forth (BaF) prediction, developed for deep feature tensors, which allows us to dramatically reduce tensor size and improve its compressibility. Our experiments with a state-of-the-art object detector demonstrate that the proposed method allows us to significantly reduce the number of bits needed for compressing feature tensors extracted from deep within the model, with negligible degradation of the detection performance and without requiring any retraining of the network weights.
We achieve a 62\% and 75\% reduction in tensor size while keeping the loss in accuracy of the network to less than 1\% and 2\%, respectively.

\end{abstract}

\vspace{-.1cm}
\section{Introduction}
With the stunning success of deep neural networks (DNN) over the last several years, AI-enabled devices have been present in practice for a multitude of application scenarios~\cite{poniszewska2018endowing}. One straightforward realization is to operate the entire neural network on a mobile device (mobile-only approach), but it can consume available resources quickly. The most common approach nowadays is to use the mobile device as a sensor to acquire the data, then transfer it to the cloud in order to run a complex DNN (cloud-only approach). However, this approach can cause congestion problems due to the growth of the volume of data transmitted over the network and the number of devices linked to the cloud. To address this problem, recent studies in collaborative intelligence have developed optimized deployment strategies~\cite{kang2017neurosurgeon, jointdnn, dfc_for_collab_object_detection, Choi2018NearLosslessDF, eshratifar2019bottlenet, eshratifar2019towards}.

The key idea of the collaborative intelligence is to split a DNN such that the computational workload between the mobile device and the cloud is optimized in terms of latency and energy consumption. Furthermore, considering the impact of the volume of data on the congestion problem, it is desirable to compress and transfer a lesser volume of data to the cloud, unless the inference performance loss is large~\cite{jointdnn}. Previous studies~\cite{dfc_for_collab_object_detection, Choi2018NearLosslessDF} have explored the efficacy of compressing deep feature tensors using conventional standard codecs, in the context of object detection and image classification. Also,~\cite{eshratifar2019bottlenet} suggests a method to first reduce the dimensionality of the deep feature tensor, then compress it using a codec.

In this paper, we propose an alternative method for dimensionality reduction of deep feature tensors produced by 
a state-of-the-art object detector. 
Our approach involves selecting a subset of tensor channels from which other channels can be predicted, compressing only this subset, and restoring the whole tensor in the cloud with negligible loss to the object detection accuracy.
The key contributions of this method are: 
\begin{itemize}
    \setlength{\itemsep}{1pt}
    \setlength{\parskip}{0pt}
    \setlength{\parsep}{0pt}
    \item We propose a deep feature tensor compression method, in which a selected subset of channels of the tensor 
    are quantized, compressed, and then transferred to the cloud.
    \item We introduce a novel back-and-forth prediction method to restore the original tensor from the compressed sub-tensor.
\end{itemize}
We briefly review related work in Section~\ref{sec:prior_work}, and in Section~\ref{sec:proposed_method} we present the details of the proposed method. Experimental results are presented in Section~\ref{sec:experimental_results}, followed by conclusions in Section~\ref{sec:conclusions}.

\vspace{-.1cm}
\section{Prior work}
\label{sec:prior_work}
Recent studies~\cite{jointdnn, dfc_for_collab_object_detection, Choi2018NearLosslessDF, eshratifar2019bottlenet} have discussed various methods to compress deep feature tensors transmitted to the cloud. In~\cite{jointdnn}, a feature map quantized to 8 bits was compressed by PNG, which is a lossless compression tool, so that the degradation of inference accuracy was negligible. However, the compression ratio was limited. Choi~\textit{et al.}~\cite{dfc_for_collab_object_detection} studied lossy coding of compressed deep features in the context of high-accuracy object detection~\cite{YOLO2}. For lossy coding, HEVC~\cite{hevc_std} was employed, and it achieved up to a 70\% reduction in coded tensor size, while preserving the same mean Average Precision (mAP). For lossless coding,~\cite{Choi2018NearLosslessDF} customized a lossless codec based on a statistical analysis of deep features from different DNNs and demonstrated that the proposed lossless tool on average marginally outperforms the latest lossless codecs such as HEVC and VP9 for four different DNNs. More recently,~\cite{eshratifar2019towards} inserted an extra autoencoder between the split sub-networks such that further dimension reduction and restoration of tensor data are conducted before and after JPEG compression, respectively. Furthermore, a training strategy referred to as compression-aware training was introduced to fully incorporate the inserted block into the target DNN. After end-to-end training from scratch, an 80\% reduction in bits with less than 2\% accuracy drop compared to the cloud-only approach is reported in the context of image classification. However, it is not clear how accurately the inserted block restores the input tensor. Moreover, this approach necessitates end-to-end training to gain the reported performance. In contrast, our work straightforwardly focuses on restoration of the deep feature data with the proposed tensor compression, which does not require end-to-end retraining. In the next section, we present how our proposed method predicts the original tensor using the compressed data as an input of a small trainable block along with the pre-trained weights of the given network such that the loss of inference accuracy is minimized without requiring end-to-end training.

\section{Proposed method}
\label{sec:proposed_method}

\begin{figure}[t]
    \begin{minipage}[b]{1.0\linewidth}
    \centering
    \includegraphics[width=\textwidth]{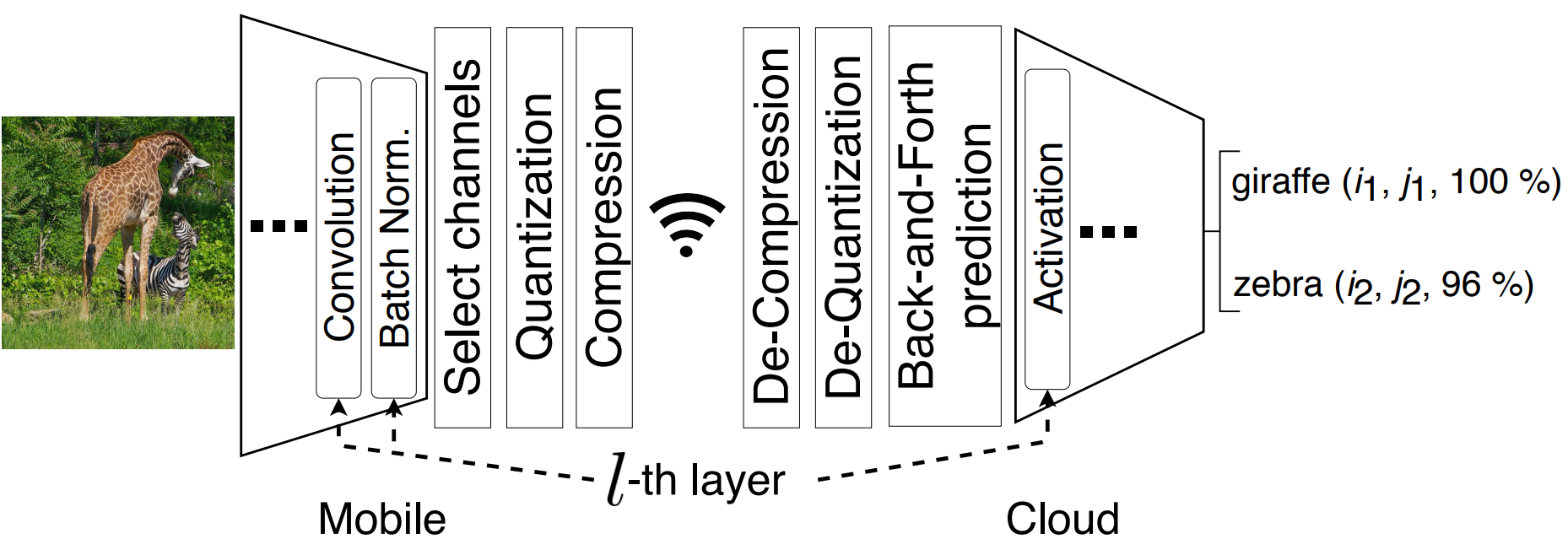}
    \end{minipage}
\caption{Proposed method of the deep feature compression in the context of the object detection.  }
\label{fig:proposed_method}
\vspace{-.3cm}
\end{figure}

The components making a convolutional neural network include convolution, activation ($\sigma$) and/or batch normalization ($\textup{BN}$)~\cite{ioffe2015batch}. The relationship between the input and the output at the $l$-th layer in a network is 
\begin{equation}
\mathbf{Y}_p^{(l)} = \sigma\left( \textup{BN}\left(\sum_{q=0}^{Q} \mathbf{W}_{p,q}^{(l)}\ast \mathbf{X}_q^{(l)}\right)\right)
\label{eq:split_layer}
\vspace{-.1cm}
\end{equation}
\noindent where 
$\mathbf{W}_{p,q}^{(l)} \in \mathbb{R}^{L \times L}$ is the trained filter associated with the $q$-th input channel $\mathbf{X}_q^{(l)}$ for the $p$-th output channel $\mathbf{Y}_p^{(l)}$. Symbol $\sum$ denotes the matrix summation of outputs of the convolution operation $\ast$ across channels. $\textup{BN}(\cdot)$ is a linear function, but $\sigma(\cdot)$ is typically a non-linear function. Previous studies suggest methods that compress the output tensors of the activation function. 
In contrast, we propose to split a network before the activation function within the $l$\nobreakdash-th layer. 

Specifically, this paper explores the proposed method with a state-of-the-art object detection model, YOLO version 3 (YOLO-v3)~\cite{redmon2018yolov3}. Due to the complex structures of this network such as residual connections and multi-scale detection, there are not many candidate layers at which to cut, when considering the volume of tensor data and computational complexity.
One good candidate is layer $l=12$, comprising convolutions with a stride of 2 and filter size $L=3$, followed by $\textup{BN}$ and $\sigma$, because not only the data volume to compress is smallest among other candidates, but also residual connections do not pass around that layer.

\begin{figure}[t]
    \centering
    \begin{minipage}[b]{0.95\linewidth}
    \centering
    \includegraphics[width=\textwidth]{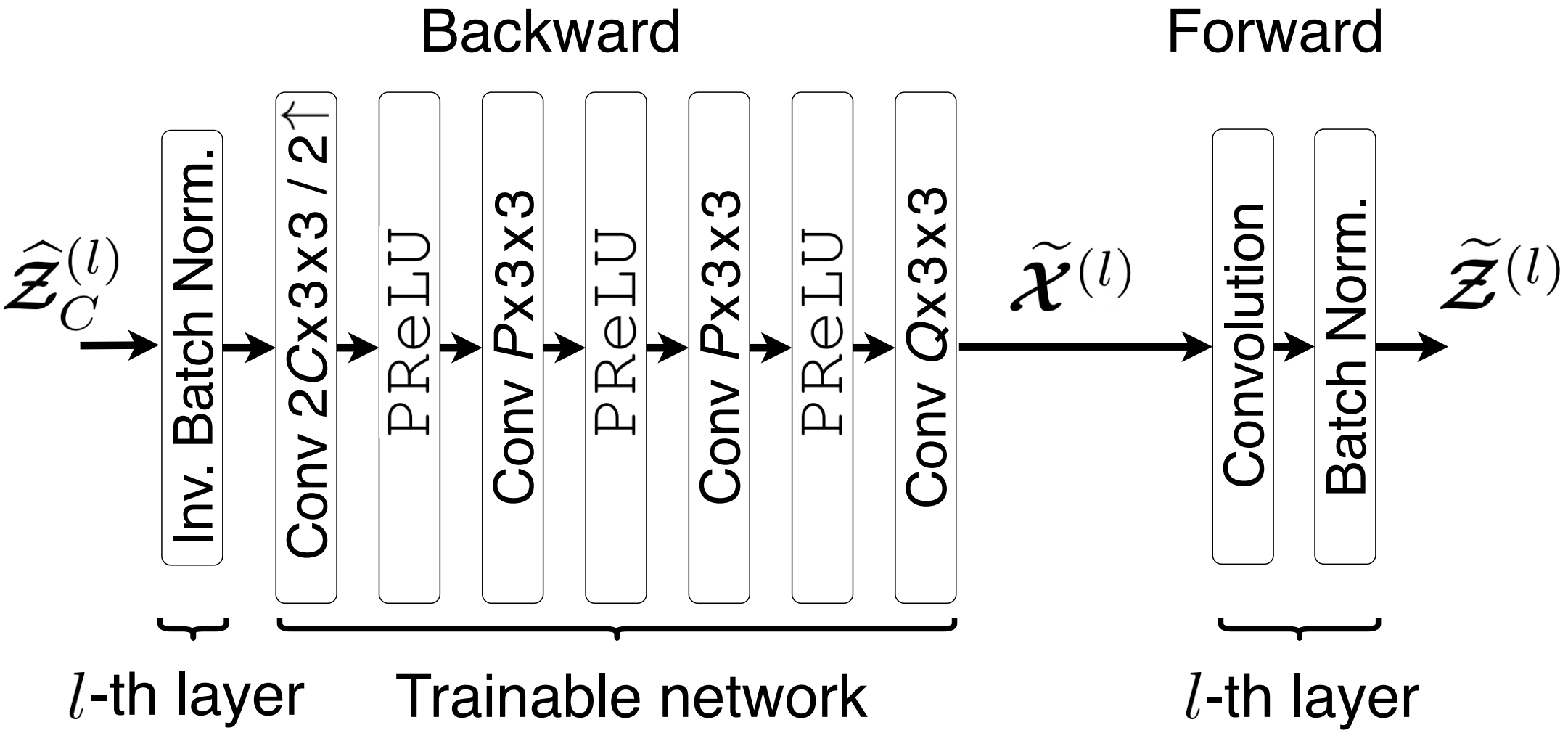}
    \end{minipage}
\caption{Structure of the proposed back-and-forth (BaF) predictor, deployed in the cloud.}
\label{fig:proposed_BaF}
\vspace{-.4cm}
\end{figure}

As shown in Fig.~\ref{fig:proposed_method}, the last computation on the mobile device and the first computation in the cloud are $\textup{BN}$ and $\sigma$ of the $l$-th layer, respectively. Let $\mathbf{Z}_p^{(l)}$
be the $p$-th channel of the output of $\textup{BN}$, and let $\mathbfcal{Z}^{(l)} = [\mathbf{Z}_1^{(l)}, ..., \mathbf{Z}_P^{(l)}]$ be the corresponding tensor with all $P$ channels. After the $\textup{BN}$ computation, 
we select a subset of $C$ channels, 
$C < P$, from the $\textup{BN}$ output based on pre-computed statistics, as described below. Let the resulting tensor with $C$ channels be denoted $\mathbfcal{Z}_C^{(l)}$.  We quantize and compress $\mathbfcal{Z}_C^{(l)}$ for transfer to the cloud. 
In the cloud, the compressed tensor is reconstructed to $\widehat{\mathbfcal{Z}}_C^{(l)}$. 
In order to restore the complete 
tensor with $P$ channels, we employ a small trainable network block as shown in Fig.~\ref{fig:proposed_BaF}, referred to as Back-and-Forth (BaF) prediction, using $\widehat{\mathbfcal{Z}}_C^{(l)}$ as input. The trainable network computes a prediction $\widetilde{\mathbfcal{X}}^{(l)} = [\widetilde{\mathbf{X}}_1^{(l)}, ..., \widetilde{\mathbf{X}}_Q^{(l)}]$ of all the \emph{input} channels 
to layer $l$, i.e. a backward prediction, then the $l$-th layer filters and $\textup{BN}$ are applied to  $\widetilde{\mathbfcal{X}}^{(l)}$ 
to generate a forward prediction of all channels of the $\textup{BN}$ output, $\widetilde{\mathbfcal{Z}}^{(l)}$. Finally, the generated 
tensor goes through the activation function in the $l$-th layer and through the remaining layers 
of the deep network. 


\subsection{Channel selection}
In order to reduce the volume of tensor data to compress, we first select $C$ out of $P$ channels of the $\textup{BN}$ output at layer~$l$. 
Note that this selection process does not add to the complexity of network because indices of selected channels 
are pre-determined by offline analysis. 
Since the selected channels will be used to predict the input of the $l$-th layer, $\mathbf{X}_q^{(l)}$, we choose the $\textup{BN}$ output channels $\mathbf{Z}_p^{(l)}$ that are the most correlated with the input channels $\mathbf{X}_q^{(l)}$. Note that, due to a stride of 2, the size of $\mathbf{X}_q^{(l)}$ is four times that of $\mathbf{Z}_p^{(l)}$. Therefore, by choosing different offsets, we generate four downsampled versions of $\mathbf{X}_q^{(l)}$ indexed by $s=0,1,2,3$. Let $\mathbf{x}_{q,s}$ be the vectorized $s$-th downsampled version of $\mathbf{X}_q^{(l)}$ and let $\mathbf{z}_p$ be the vectorized version of $\mathbf{Z}_p^{(l)}$. We compute absolute pairwise correlation coefficients
by 
\begin{equation}
\rho_{p,q} = \frac{1}{4}\sum_{s=0}^{3}{\left | \frac{(\mathbf{z}_p - \bar{z}_p) \cdot (\mathbf{x}_{q,s} - \bar{x}_{q,s})}{\left \| (\mathbf{z}_p - \bar{z}_p)\right \|_2 \left \| (\mathbf{x}_{q,s} - \bar{x}_{q,s}) \right \|_2} \right | },
\label{eq:correlation_distance}
\end{equation}
where $\bar{z}_p$ and $\bar{x}_{q,s}$ represent the mean values of $\mathbf{z}_p$ and  $\mathbf{x}_{q,s}$, respectively. 
So $\rho_{p,q}$ is the average of the absolute correlation coefficients between $\mathbf{Z}_p^{(l)}$ and the four downsampled versions of $\mathbf{X}_q^{(l)}$. 
The index of the channel $\mathbf{Z}_p^{(l)}$ that has the highest total correlation with respect to all $\mathbf{X}_q^{(l)}$ is found as
\begin{equation}
p^{*} = \argmax_{p}\sum_{q=0}^{k}{\rho_{p,q}}\,.
\label{eq:list_establish}
\end{equation}
We repeat this selection process with the remaining channels until all $C$ channels have been processed, resulting in an ordered list of $C$ channels $\mathbf{Z}_p^{(l)}$ based on decreasing correlation with all $\mathbf{X}_q^{(l)}$.  
To compute the coefficients for a given pre-trained network, we use the network weights up to the $l$-th layer, along with a randomly selected set of 1k images from the 2014 COCO training dataset~\cite{lin2014microsoft} as input to the network.

\subsection{Quantization and tiling}
In order to compress the selected channels, 
we first apply $n$-bit uniform scalar quantization to each channel separately and cast floats to integers: 
\begin{equation} 
\texttt{Q}\left(\mathbf{Z}_p^{(l)}\right) = \textup{round} \left (\frac{\mathbf{Z}_p^{(l)}-m_p^{(l)}}{M_p^{(l)}-m_p^{(l)}}\cdot (2^{n}-1)\right )
\label{eq:quantized_mat}
\end{equation}
\noindent where $\texttt{Q}\left(\mathbf{Z}_p^{(l)}\right)$ is the quantized feature channel 
and $m_p^{(l)}$ and $M_p^{(l)}$ are the minimum and maximum of $\mathbf{Z}_p^{(l)}$  
rounded to 16-bit floating point precision. 
$m_p^{(l)}$ and $M_p^{(l)}$ 
are transmitted to the cloud as side information, and the extra $C\cdot 32$ bits needed for these values is included in the total bit counts in our experiments. The quantized channels are rearranged into a tiled image for compression using a conventional image codec. To make the tiled image rectangular in shape, the number of channels arranged over the width and height of the tiled image are $\textup{ceil}\left(\frac{1}{2}\log_{2}C\right)$ and $\textup{floor}\left(\frac{1}{2}\log_{2}C\right)$, respectively. Here, ceil$(\cdot)$ and floor$(\cdot)$ represent ceiling and flooring to the nearest integer. To avoid any empty areas in the tiled image, we always choose $C$ to be a power of 2. 

\subsection{Back-and-forth prediction}
When the compressed bitstream arrives in the cloud, the tiled image is decompressed and rearranged back into the quantized set of sub-channels. Next, inverse quantization
is performed:
\begin{equation}
\widehat{\mathbf{Z}}_p^{(l)} = \frac{\texttt{Q}\left(\mathbf{Z}_p^{(l)}\right)}{2^{n}-1}\cdot\left(M_p^{(l)}-m_p^{(l)}\right) + m_p^{(l)}.
\label{eq:invQ}
\end{equation}
Because 
only a subset of $C$ of the original $P$ channels were transmitted, 
it is necessary to restore the other channels 
to complete the inference task in the cloud. 
This is accomplished using a trainable network shown in Fig.~\ref{fig:proposed_BaF}. 

Our approach consists of two processes: backward prediction and forward prediction, hence the name Back-and-Forth (BaF) prediction. The beginning of the backward process is to do inverse 
$\textup{BN}$, and then the trainable network in Fig.~\ref{fig:proposed_BaF} essentially needs to perform deconvolution from a limited number of channels. 
The deconvolution network consists of four convolutional layers having 3$\times$3 kernels followed by \texttt{PReLU} activations, except for the last convolutional layer which has identity activation. The output of the deconvolution network is an estimate of all \emph{input}  channels of the $l$-th layer, $\widetilde{\mathbfcal{X}}^{(l)}$. Due to the resolution differences between $\widehat{\mathbfcal{Z}}_C^{(l)}$ and $\widetilde{\mathbfcal{X}}^{(l)}$, the first convolutional layer employs an up-sampling operation which increases width and height of the channels by 2. Once $\widetilde{\mathbfcal{X}}^{(l)}$ is obtained, the forward predictor is simply the $l$-th layer convolution and $\textup{BN}$, both with pre-trained weights, which produces an estimate of all $P$ channels of the $\textup{BN}$ output, 
$\widetilde{\mathbfcal{Z}}^{(l)}$. 

Since the BaF predictor generates an estimate of all channels of the $\textup{BN}$ output of layer $l$, it not only generates estimates of non-transmitted channels, but it also generates estimates of the transmitted $C$ channels. Hence, for each of the transmitted $C$ channels, we now have two possible choices: $\widetilde{\mathbf{Z}}_p^{(l)}$, which is generated through BaF prediction, and $\widehat{\mathbf{Z}}_p^{(l)}$, which is the result of inverse quantization~(\ref{eq:invQ}), for $p<C$. 
For these $C$ channels, we select the final value of the element $(i,j)$ in channel $p$ as:
\begin{equation} 
\widetilde{\mathbf{Z}}_p^{(l)}(i,j) \leftarrow 
\begin{cases}
    \begin{array}{l@{\hskip4pt}l} 
        \textstyle{\widetilde{\mathbf{Z}}_p^{(l)}(i,j),} & \textstyle{\text{if } \texttt{Q}\left(\widetilde{\mathbf{Z}}_p^{(l)}(i,j)\right)= \texttt{Q}\left(\widehat{\mathbf{Z}}_p^{(l)}(i,j)\right)} \\
        \textstyle{b},& \textstyle{\text{otherwise}}
    \end{array}
\end{cases}
\label{eq:consolidation}
\vspace{2pt} 
\end{equation}
\noindent where $b$ is the inverse quantized boundary value of the quantizer bin $\texttt{Q}\left(\widehat{\mathbf{Z}}_p^{(l)}(i,j)\right)$ that is closest to $\widetilde{\mathbf{Z}}_p^{(l)}(i,j)$. In other words, if $\widetilde{\mathbf{Z}}_p^{(l)}(i,j)$ is consistent with quantization, i.e. $\widehat{\mathbf{Z}}_p^{(l)}(i,j)$ and $\widetilde{\mathbf{Z}}_p^{(l)}(i,j)$ fall into the same quantizer bin (first case in~(\ref{eq:consolidation})), then $\widetilde{\mathbf{Z}}_p^{(l)}(i,j)$ is taken as final value. Otherwise, if $\widetilde{\mathbf{Z}}_p^{(l)}(i,j)$ falls in a different quantizer bin compared to $\widehat{\mathbf{Z}}_p^{(l)}(i,j)$, 
$b$ is taken as the final value. This way, the distance of the final reconstructed value from $\widetilde{\mathbf{Z}}_p^{(l)}(i,j)$ is minimized while satisfying the quantization condition.  
Lastly, the end goal of the DNN, e.g. object detection, is completed by passing the full reconstructed tensor to the remaining layers of the DNN in the cloud.

\section{Experiments}
\label{sec:experimental_results}
\vspace{-.2cm}
We demonstrate our proposed method in the context of a recent object detection network, YOLO-v3. For this network, we use the weights provided in~\cite{darknet}, which were pre-trained using the 2014 COCO dataset~\cite{lin2014microsoft}. Additionally, our proposed BaF prediction network has been trained on the same training dataset, but only for the randomly selected 2M images. To obtain the input to the BaF model, we pass the samples through the first part of the network and save the outputs of $\textup{BN}$ in layer $l=12$ as files. Because the input resolution of the YOLO-v3 network is $512\times512$, the dimension of the output tensor $(N\times M \times P)$ at the 12th layer is $64\times64\times256$. The loss function, referred to as Charbonnier penalty function~\cite{charbonnier1994two}, used to optimize the BaF prediction is defined by
\begin{equation} 
\mathcal{L} = \sum \sqrt{(\mathbfcal{Y}^{(l)} - \sigma(\widetilde{\mathbfcal{Z}}^{(l)}))^2 + \epsilon^2} 
\label{eq:loss_function}
\vspace{-.1cm}
\end{equation}
\noindent where the sum ($\sum$) accumulates all elements over three dimensions, and the regularization constant $\epsilon = 10^{-3}$. The consolidation function of~(\ref{eq:consolidation})
is ignored while training. We trained the BaF network on a GeForce GTX 1080 GPU with 12GB memory over 7.5M iterations. 

\begin{figure}[t]
    \begin{minipage}[b]{1.0\linewidth}
    \centering
    \includegraphics[width=\textwidth]{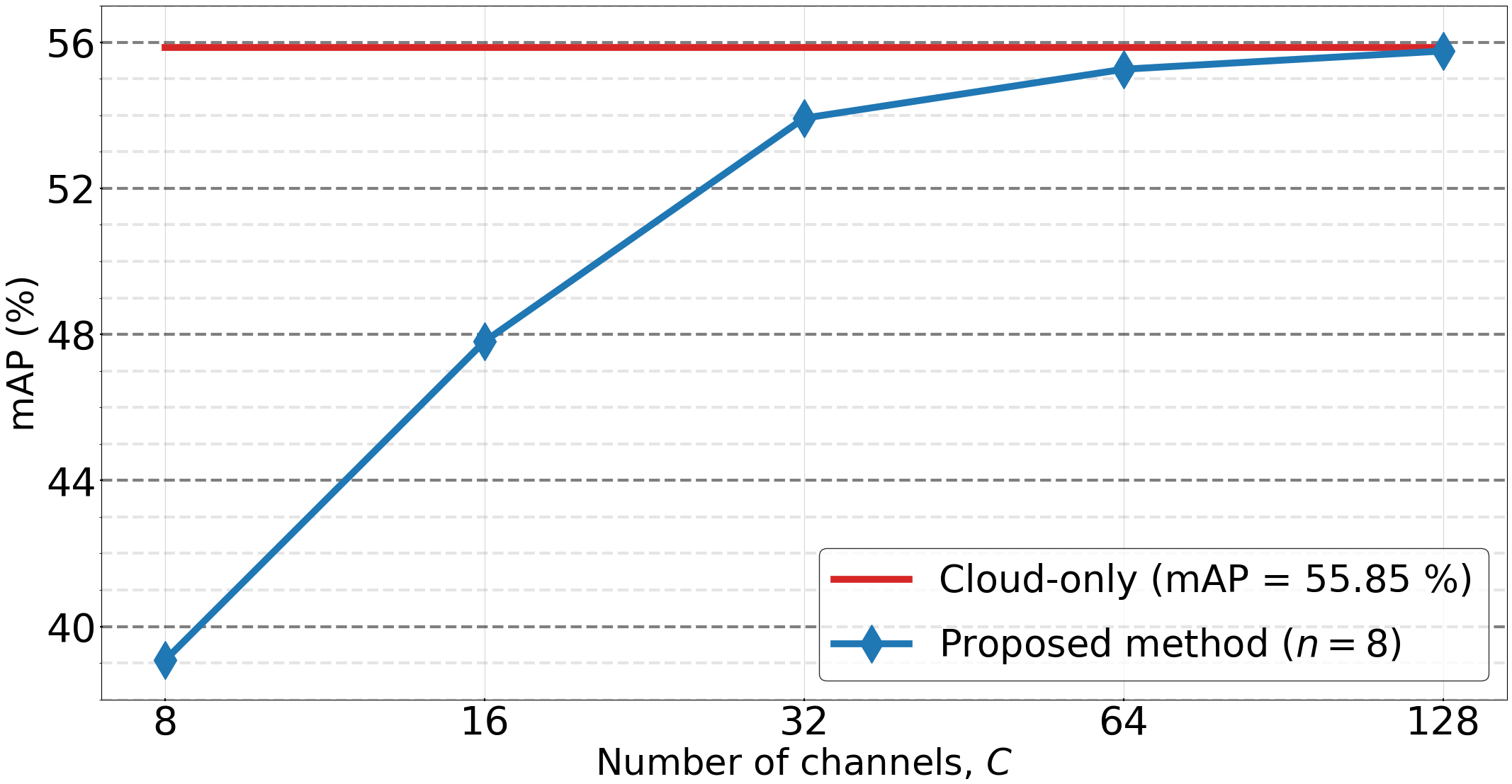}
    \end{minipage}
\caption{mAP performance curve vs. number of channels with $n=8$ bits, compared to the cloud-only approach}
\label{fig:performance_regarding_different_number_of_channels}
\vspace{-.3cm}
\end{figure}

To evaluate the prediction performance of the proposed method, separate BaF models for $C =\{8, 16, 32, 64, 128\}$ channels and $n=8$ bits are trained. The tensors predicted by these models are input to the remaining object detection sub-network in the cloud in order to obtain
mAP values.
For testing, about 5k images from the 2014 COCO validation dataset (separate from the training dataset) are used. Fig.~\ref{fig:performance_regarding_different_number_of_channels} presents
mAP
with respect to the number of channels. The benchmark unmodified YOLO-v3 mAP is 55.85\%, as indicated in red.
With
128
channels, there is no loss in mAP.
With
64
channels, the mAP is $55.26\%$, which is a degradation of less than 1\% compared to the benchmark. Hence, a good balance between coding efficiency and near-lossless mAP performance is achieved using the proposed method with only a quarter of the channels from the complete tensor. 

\begin{figure}[t]
    \begin{minipage}[b]{1.0\linewidth}
    \centering
    \includegraphics[width=\textwidth]{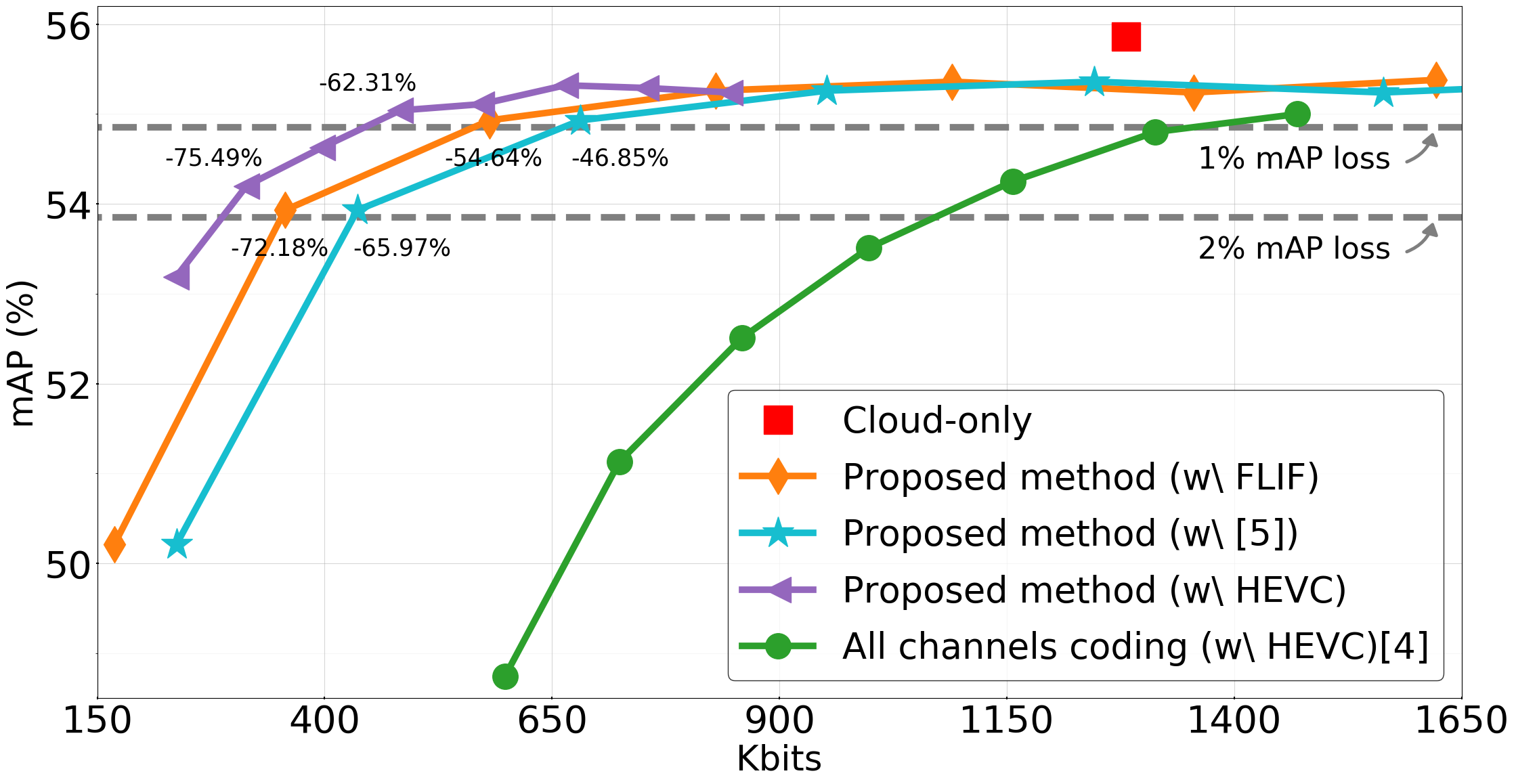}
    \end{minipage}
\caption{Performance comparison against cloud-only approach}
\label{fig:performance_regarding_different_nbits}
\vspace{-.3cm}
\end{figure}
In Fig.~\ref{fig:performance_regarding_different_nbits}, we demonstrate the impact of quantization with $n=\{2, 3, 4,..,8\}$ on the proposed method when $C = 64$. To compress the tiled quantized tensors, we use the lossless codec FLIF~\cite{sneyers2016flif}, as it adaptively supports low precision samples, or we use the
lossless tensor compression method of ~\cite{Choi2018NearLosslessDF}.
We individually train BaF models with respect to $n$ and compare the performance against the cloud-only, i.e. unmodified YOLO-v3 approach.
For comparison,
we use the method of~\cite{dfc_for_collab_object_detection} to compress
all channels quantized with 8-bit, using HEVC with different quantization (QP) parameters.
The percentage beside some data points in the plot denotes the compression ratio compared to the cloud-only approach, i.e. the reduction in file size of the compressed tensor vs. the compressed image input to an unmodified network. Therefore, we achieve about 55\% and 72\% in bits savings by allowing less than 1\% and 2\% mAP loss, respectively. In terms of BD-Bitrate-mAP~\cite{dfc_for_collab_object_detection}, the proposed method saves over 90\% compared to the case compressing all channels using HEVC. Additionally, we further compress the 6-bit version of the quantized tensor using HEVC, as demonstrated by the purple curve. 
We achieve 3--8\% extra improvement for the same mAP as compared to using FLIF, resulting in an overall 62\% and 75\% bit savings for less than 1\% and 2\% mAP loss, respectively.
Moreover, this approach even outperforms the case when JPEG images are transcoded by HEVC after color-sub sampling, by about 1--2\% in terms BD-Bitrate-mAP.

\vspace{-.3cm}
\section{Conclusion}
\vspace{-.2cm}
\label{sec:conclusions}
We proposed a novel back-and-forth (BaF) neural network that uses a compressed set of tensor sub-channels to predict the tensor to be used as the input to the remaining network in the cloud for collaborative intelligence applications, without requiring any re-training of the original DNN. By pre-computing the correlation between input and output channels of the last convolutional operation performed in the first part of the network, e.g. in a mobile device, we significantly reduce the degree of the tensor dimension needed to be compressed, while preserving the object detection performance.
We also presented a method for selecting predictions from either the BaF prediction network or from the decoded sub-channels in order to reduce deviation from the original sub-channels.
When incorporated into the YOLO-v3 object detection network, we
achieved a 62\% and 75\% reduction in tensor size
with less than 1\% and 2\% loss in mAP, respectively.


\bibliographystyle{IEEEbib}
\bibliography{ref}

\begin{thebibliography}{10}

\bibitem{poniszewska2018endowing}
A.~Poniszewska-Maranda, D.~Kaczmarek, N.~Kryvinska, and F.~Xhafa,
\newblock ``Endowing iot devices with intelligent services,''
\newblock in {\em Proc. Int. Conf. Emerging Internetworking, Data \& Web
  Technol.}, 2018, pp. 359--370.

\bibitem{kang2017neurosurgeon}
Y.~Kang, J.~Hauswald, C.~Gao, A.~Rovinski, T.~Mudge, J.~Mars, and L.~Tang,
\newblock ``Neurosurgeon: Collaborative intelligence between the cloud and
  mobile edge,''
\newblock in {\em Proc. 22nd ACM Int. Conf. Arch. Support Programming Languages
  and Operating Syst.}, 2017, pp. 615--629.

\bibitem{jointdnn}
A.~E. Eshratifar, M.~S. Abrishami, and M.~Pedram,
\newblock ``{JointDNN}: an efficient training and inference engine for
  intelligent mobile cloud computing services,''
\newblock {\em arXiv preprint arXiv:1801.08618}, 2018.

\bibitem{dfc_for_collab_object_detection}
H.~Choi and I.~V. Baji\'{c},
\newblock ``Deep feature compression for collaborative object detection,''
\newblock in {\em Proc. IEEE ICIP'18}, 2018.

\bibitem{Choi2018NearLosslessDF}
H.~Choi and I.~V. Baji\'{c},
\newblock ``Near-lossless deep feature compression for collaborative
  intelligence,''
\newblock {\em 2018 IEEE 20th International Workshop on Multimedia Signal
  Processing (MMSP)}, pp. 1--6, 2018.

\bibitem{eshratifar2019bottlenet}
A.~E. Eshratifar, A.~Esmaili, and M.~Pedram,
\newblock ``Bottlenet: A deep learning architecture for intelligent mobile
  cloud computing services,''
\newblock {\em arXiv preprint arXiv:1902.01000}, 2019.

\bibitem{eshratifar2019towards}
A.~E. Eshratifar, A.~Esmaili, and M.~Pedram,
\newblock ``Towards collaborative intelligence friendly architectures for deep
  learning,''
\newblock in {\em 20th International Symposium on Quality Electronic Design
  (ISQED)}. IEEE, 2019.

\bibitem{YOLO2}
J.~Redmon and A.~Farhadi,
\newblock ``{YOLO9000:} better, faster, stronger,''
\newblock in {\em Proc. IEEE CVPR'17}, Jul. 2017, pp. 6517--6525.

\bibitem{hevc_std}
Int. Telecommun. Union-Telecommun. (ITU-T) and Int. Standards
  Org./Int/Electrotech. Commun. (ISO/IEC~JTC 1),
\newblock ``High efficiency video coding,'' Rec. ITU-T H.265 and ISO/IEC
  23008-2, 2013.

\bibitem{ioffe2015batch}
S.~Ioffe and C.~Szegedy,
\newblock ``Batch normalization: Accelerating deep network training by reducing
  internal covariate shift,''
\newblock {\em arXiv preprint arXiv:1502.03167}, 2015.

\bibitem{redmon2018yolov3}
J.~Redmon and A.~Farhadi,
\newblock ``Yolov3: An incremental improvement,''
\newblock {\em arXiv preprint arXiv:1804.02767}, 2018.

\bibitem{lin2014microsoft}
T.-Y. Lin, M.~Maire, S.~Belongie, J.~Hays, P.~Perona, D.~Ramanan,
  P.~Doll{\'a}r, and C.~L. Zitnick,
\newblock ``Microsoft coco: Common objects in context,''
\newblock in {\em European conference on computer vision}. Springer, 2014, pp.
  740--755.

\bibitem{darknet}
J.~Redmon,
\newblock ``{Darknet: Open source neural networks in C.},''
  http://pjreddie.com/darknet/, 2013-2017,
\newblock Accessed: 2017-10-19.

\bibitem{charbonnier1994two}
P.~Charbonnier, L.~Blanc-Feraud, G.~Aubert, and M.~Barlaud,
\newblock ``Two deterministic half-quadratic regularization algorithms for
  computed imaging,''
\newblock in {\em Proc. IEEE ICIP'94}, 1994.

\bibitem{sneyers2016flif}
J.~Sneyers and P.~Wuille,
\newblock ``{FLIF}: Free lossless image format based on {MANIAC} compression,''
\newblock in {\em Proc. IEEE ICIP'16}. IEEE, 2016.

\end{thebibliography}

\end{document}